# Novel Development of LLM Driven mCODE Data Model for Improved Clinical Trial Matching to Enable Standardization and Interoperability in Oncology Research

Aarsh Shekhar
McNeil High School

Mincheol Kim
University of Chicago

*Abstract:* Each year, the lack of efficient data standardization and interoperability in cancer care contributes to the severe lack of timely and effective diagnosis, while constantly adding to the burden of cost with cancer costs nationally reaching over $208 billion in 2023 alone. Traditional methods regarding clinical trial enrollment and clinical care in oncology are often manual, time-consuming, and lack a data-driven approach. This paper presents a novel framework to streamline standardization, interoperability and exchange of cancer domains and enhance the integration of oncology-based EHRs across disparate healthcare systems. This paper utilizes advanced LLMs and Computer Engineering to streamline cancer clinical trials and discovery. By utilizing FHIR's resource-based approach and LLM-generated mCODE profiles, we ensure timely, accurate and efficient sharing of patient information across disparate healthcare systems.

Our methodology involves transforming unstructured patient treatment data, PDF's, free-text information, and progress notes into enriched mCODE profiles, facilitating seamless integration with our novel AI and ML based clinical trial matching engine. The results of this study show a significant improvement in data standardization, with accuracy rates of our trained LLM peaking at over 92% with datasets consisting of thousands of patient's data. Additionally, our LLM demonstrated an accuracy rate of 87% for SNOMED-CT, 90% for LOINC, and 84% for RxNorm codes. This trumps the current status quo, with LLM's such as GPT-4 and Claude's 3.5 peaking at an average of 77%. This paper successfully underscores the potential of our standardization and interoperability framework, paving the way for more efficient and personalized cancer treatment.

## Background Information and Introduction

### Prevalence of Data Standardization and the Onset of FHIR and LLM's in Healthcare

The healthcare industry has witnessed a significant transformation in recent years, driven by the integration of advanced technologies such as Large Language Models (LLMs) and sophisticated data standardization and analysis techniques. This shift has been particularly impactful in oncology and in addressing the unique challenges faced by rural and underserved healthcare facilities and regions. The advent and adoption of Electronic Health Records (EHRs) has laid the foundation for this digital revolution, enabling the collection and analysis of vast amounts of patient data

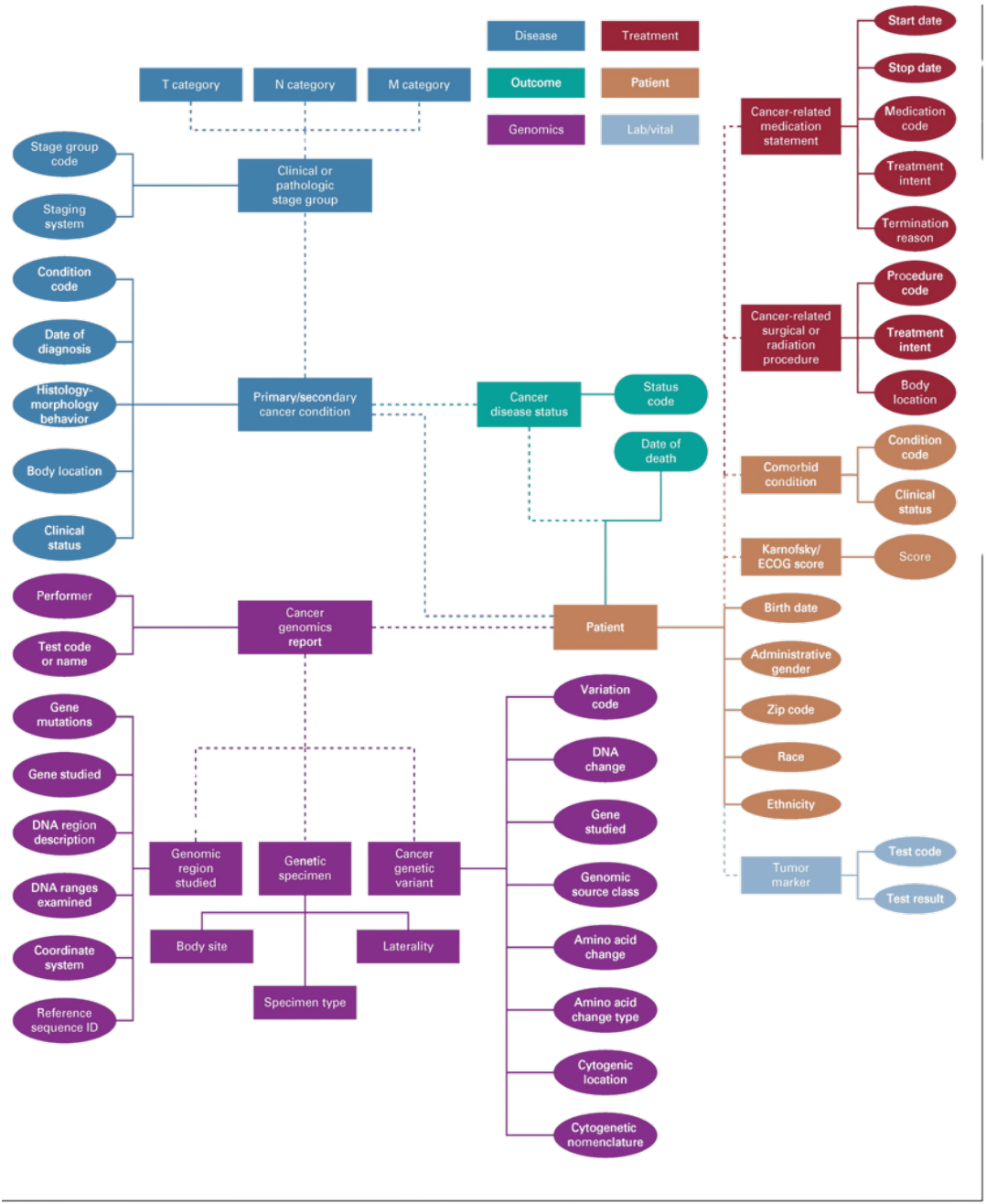

*Figure 1. Minimal Common Oncology Data Elements(mCODE), high-level conceptual model breaking down various domains, profiles, and properties [5]*

at an unprecedented scale.

EHRs (Electronic Health Records) have been a critical first step in getting all healthcare information on computers, with adoption rates in the United States reaching 96% with the help of the Government pushing the use of the HITECH Act and the Cures Act.

This recognition has been taken up by most major technology companies about the potential of healthcare data analytics, followed by a significant investment in the area. For example, Watson for Oncology by IBM attempted to leverage Artificial Intelligence (AI) to help in decision-making regarding the treatment of cancer [1]. Similarly, Apple's Health Records feature has tried to empower patients with access to information in their medical records across various healthcare providers [2]. These initiatives are laudable but again point towards the conundrum unsolved of integrating diverse data sources and translating complex medical information into actionable insights.

The introduction of LLMs into healthcare has the potential to be groundbreaking in the interpretation and analysis of data. These models process and generate text that is applicable to a human-like extent, thus providing unprecedented opportunities for deriving meaningful insights from unstructured medical data that constitute clinical notes and patient profiles. In the oncology field, LLMs show potential interest in tumor classification, treatment recommendation, and prognosis prediction [3]. However, all these sophisticated tools rely on the quality and interoperability of the underlying data. This is where the Fast Healthcare Interoperability Resources (FHIR) standard comes into play.

Of importance to this note is that FHIR finally begins to offer solutions to one of the most significant obstacles health systems have struggled with up until now - the ability of one system to communicate with another using open internet technologies and data formats.

FHIR provides a standardized approach to the sharing of healthcare data, allowing heterogeneous systems to interoperate with one another properly. In this manner, it provides a common language for healthcare data, and hence, merging the diverse sources of data is made more accessible to therefore be used in creating a more comprehensive view of patient health.

**FHIR x mCODE in the Oncology Industry**

It is in this complex context of Oncology that the combination of Fast Healthcare Interoperability Resources (FHIR) and Minimal Common Oncology Data Elements (mCODE) will become a driving force for a new age in data interoperability and analysis.

Incorporating FHIR in Oncological settings will facilitate an easy integration of many different data sources, from EHR systems to genomic databases and imaging systems. Mandel et al. (2016) further presented evidence that using applications based on FHIR leads to a significant advancement of clinical workflow and considerably improves data access in oncology settings, which directly yields a high level of clinical decisions and personalized treatment plans [4].

Even though FHIR provides this kind of infrastructure, standardization should be specified for oncology data by mCODE, hence the prerogative of this paper to begin the development and integration of this model. mCODE defines a core set of structured data elements for oncology EHRs. The standard spans six principal domains: patient, disease, lab/vital, genomics, treatment, and outcome. The patient domain involves demographics and comorbidities, while the disease domain describes a diagnosis of cancer, staging, and characteristics of the tumor. The lab/vital domain includes laboratory results and vital signs relevant to cancer care. The genomics domain is crucial for the current oncology and includes data on biomarkers and genetic alterations. The treatment domain details surgeries, radiation therapy, and systemic treatments. Finally, the outcome domain includes data on disease status, progression, and patient-reported outcomes [5].

For mCODE, each domain is made up of numerous concepts that then have data elements linked with it. For example, in the Treatment domain, there are concepts on Cancer-Related Medication Statements and Procedure Statements. These are records of specific data elements, like the intent for treatment, such as curative or palliative, start and end dates of the drug, reasons to stop medicines; and more.

The structured nature of mCODE data elements, along with FHIR interoperability capabilities, allows efficacious collection, analysis, and sharing of data across institutions and research networks – tackling the largest issue in the oncological status quo.

**The Applicability of LLM's in Healthcare Standardization**

In electronic health records (EHR), unstructured data comprises nearly 80% of the total information in a healthcare operation's data set. This is because many clinical procedures and findings, such as physician notes, radiology reports, and pathology findings, are often maintained in their original unstructured form [6]. Despite its inherent complexity, this unstructured data holds valuable information that can significantly advance cancer research and patient care. These sources provide crucial details about a patient's medical history, treatment responses, and disease progression, which are indispensable for developing personalized treatment plans and conducting research into novel therapies.

Standardized data plays a critical role in achieving consistent results and ensures the seamless integration of data from diverse sources, thereby fostering collaborative research initiatives. The challenge of converting unstructured data to a structured format is significant, especially in specialized domains like cancer, where medical terminology and concepts are highly specific. The topic of data standardization from clinical notes largely revolves around the concept of information extraction, which inherently includes term extraction, relation extraction, and the natural language processing (NLP) techniques associated with these tasks. Conventional NLP methodologies often rely on manually curated rules and human-annotated datasets for their training procedures. This results in named entity recognition (NER) or information extraction systems being less generalizable, thereby exhibiting limited portability.

Hence, this paper offers a novel framework to not only streamline this process and mitigate error, but also effectively increase the interoperability of oncological data around the world and offer the long-awaited care many constituents around the world deserve. To achieve such, we utilize LLM's, namely OpenAI's GPT 4.0, to translate free-text and unstructured information, whether that be clinical-notes or patient variables organized in a PDF/JSON format, to manually format and create a completed and compliant FHIR profile for the respective patient. With the respective FHIR profile, we deploy the LLM to break it down into the 6 respective mCODE profiles mentioned in section B. Once broken down, we integrate these 6 unique domains into an open-source framework intended for integrated trial matching, to streamline and strengthen the process of matching patients with the most effective trials to increase progressive cancer research in the status quo.

**Methods and Model Framework/Design**

This section delineates our multifaceted approach to enhancing oncology data standardization and interoperability. We explore the integration of synthetic data generation, FHIR profile implementation, and mCODE standards. Our methodology encompasses the training and guidance of Large Language Models (LLMs) for processing free-text and generating FHIR profiles, alongside the development of mechanisms for storing and categorizing these profiles within mCODE domains.

Throughout this segment, this paper effectively outlines approaches from development of patient profiles to accuracy assessment protocols using FHIR frameworks while dis-cuss all of their interpretative significance. This comprehensive framework aims to address the critical challenges in healthcare informatics identified in our introduction.

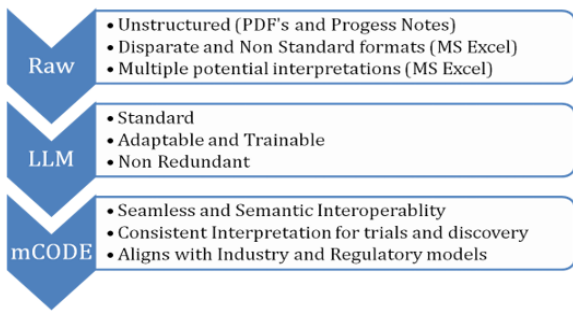

*Figure 2. Simplified Workflow of Conversion Process Between Unstructured Data and Full-Fledged mCODE Compliant Profiles*

| cancer_type_raw | histology_type_raw | histology_grade_raw | primary_tnm_raw | regional_tnm_raw | metastatic_sites_raw | laterality_raw | numerical_staging_raw |
|---|---|---|---|---|---|---|---|
| Cervical Cancer | AGGRESSIVE B-CELL LYMPHOMA | G1 (low grade; well differentiated) | pT | pN0 | Not Documented | Not Documented | IA |
| Hematologic Malignancy - Leukemia (ALL; CLL; AML; CML) | Acute promyelocytic leukemia | G1 (low grade; well differentiated) | pT1 | pN0(i+) | Lymph Node(s) - Distant | Unilateral - Left | IB |
| Not Documented | Acute promyelocytic leukemia (APL) | G2 (intermediate grade; moderately differentiated) | pT1C | pN0(sn) | nan | Unilateral - Right | II |
| Gynecological Other or Vaginal Cancer | Adenocarcinoma | G2 (intermediate grade; moderately differentiated) | pT2 | pN1 | Brain/Central Nervous System (CNS) | Unilateral - Not Docume | III |
| AML | Adenocarcinoma - Intestinal type | G3 (high grade; poorly differentiated) | pT1B | pN1a | Bone(s) | Bilateral | IIIA |

Table 1. Diagnostic Characteristics of Raw Variables Extracted From Free-Text PDF Annotations From Physicians

**Visualization of Free-Text/Unstructured Data and FHIR Profiles**

Free-text data refers to narrative text that clinician's input into electronic health records (EHRs) without any predefined structure. This form of unstructured data often includes patient histories, clinical notes, discharge summaries, and other narrative descriptions. Unstructured data is the most abundant source of unstructured healthcare industry because it provides clinicians with flexibility and ease to record detailed and nuanced patient information [7].

The conversion of free-text data into structured formats is a critical step in developing a comprehensive FHIR (Fast Healthcare Interoperability Resources) mCODE (minimal Common Oncology Data Elements) profile. This profile standardizes oncology data elements to enhance data interoperability, enabling better data sharing, analysis, and clinical decision-making. Nevertheless, the unstructured (free-text) nature of free text data presents utilization challenges due to its inherent non-standardized format that is not easily integrated across different healthcare systems.

Transforming free-text data into structured formats involves the application of natural language processing (NLP) and machine learning techniques to extract key medical concepts and translate them into standardized data elements. Hence the usage and development of our GPT-trained LLM to transform the respective free-text and unstructured data. Such structured data can be converted to conforming FHIR mCODE profile, which ensures that vital patient information is suitably defined and readily shareable across those platforms and stakeholders.

The benefits of converting unstructured free-text data into structured formats are manifold. It enhances the quality and accessibility of patient data, facilitates more effective data analytics, and improves clinical decision-making. Moreover, structured data is crucial for achieving true interoperability in the healthcare industry, as it allows for seamless data exchange, reducing redundancies and improving care coordination. By starting with the conversion of free-text data, healthcare providers can take a significant step towards achieving the comprehensive and interoperable patient records envisioned by FHIR mCODE standards.

| cancer_type_raw | histology_type_raw | histology_grade_raw | metastatic_sites_raw | laterality_raw | numerical_staging_raw |
|---|---|---|---|---|---|
| ['Not Documented'] | ['Not Documented'] | ['Not Documented'] | pred_abdomen | ['Breast cancer, female'] | {'III': True} |
| ['Multiple Myeloma'] | ['Multiple Myeloma'] | ['G3'] | pred_pelvis | ['Not Documented'] | {'IVB': True} |
| ['Colon cancer'] | ['Colon cancer'] | ['grade 3'] | pred_bones | ['Colon cancer'] | {'III': true} |
| ['Primary malignant neoplasm of female breast (disorder)'] | ['Primary malignant neoplasm of female breast (disorder)'] | ['G2'] | pred_distant_lymph_nod | ['Primary malignant neoplasm of female breast (disorder)'] | {'IIIA': True} |
| ['Sigmoid colon cancer'] | ['Sigmoid colon cancer'] | ['2: Intermediate combined histologic grade (moderately favorable)'] | pred_local_lymph_nodes | ['Sigmoid colon cancer'] | {'T2': True} |

Table 2. Diagnostic Characteristics of Raw Variables Extracted From Free-Text Clinical/Progress Notes From Physicians

**Figure 3.**
*Representational Example of Free-Text Data (Clinical Notes) from Physicians Regarding Real-Time Cancer Patients*

### Process of Training/Guiding LLM to Translate Free-Text Data Into FHIR Profiles

The translation of free-text data, including clinical notes to a FHIR (Fast Healthcare Interoperability Resources) compliant patient profile is processed in multiple steps which aim at transforming unstructured clinical information into structured and interoperable formats. The methodology, which uses complex natural language processing (NLP) algorithms in conjunction with standardized coding systems and thorough validation procedures to ensure the quality of FHIR profiles for integration within electronic health record (EHR) environments and other healthcare data exchange services [8].

The initial step in this process involves the acquisition and preparation of free-text and unstructured clinical data. This data is typically available in various formats such as PDF documents, text files, or direct inputs from clinical documentation systems. The primary objective of the preparation phase is to preprocess the data, which involves cleansing and structuring the text to facilitate efficient information extraction. This includes removing irrelevant information, such as administrative data or extraneous comments, and normalizing text by correcting typographical errors and standardizing abbreviations and acronyms. This preprocessing ensures that the focus remains on clinically relevant content, thereby enhancing the accuracy of subsequent NLP tasks.

Following data preparation, advanced natural language processing (NLP) techniques are employed to interpret and extract structured information from the free-text clinical notes. This phase is crucial as it bridges the gap between unstructured text and structured data. NLP tools, including Named Entity Recognition (NER) models, are utilized to identify and categorize key medical entities such as patient demographics, clinical conditions, medications, procedures, observations, and outcomes. These models were trained

```
{
  "resource": {
    "resourceType": "MedicationRequest",
    "id": "medicationrequest-1",
    "status": "active",
    "intent": "order",
    "medicationCodeableConcept": {
      "coding": [
        {
          "system": "http://www.nlm.nih.gov/research/umls/rxnorm",
          "code": "308056",
          "display": "Cisplatin"
        },
        {
          "system": "http://www.nlm.nih.gov/research/umls/rxnorm",
          "code": "308136",
          "display": "Pemetrexed"
        }
      ]
    },
    "subject": {
      "reference": "Patient/patient-1"
    },
    "dosageInstruction": [
      {
        "text": "Administered as part of chemotherapy regimen"
      }
    ]
  }
},
{
  "resource": {
    "resourceType": "Procedure",
    "id": "procedure-1",
    "status": "completed",
    "code": {
      "coding": [
        {
          "system": "http://snomed.info/sct",
          "code": "367336001",
          "display": "Chemotherapy"
        }
      ]
    },
    "subject": {
      "reference": "Patient/patient-1"
    },
    "performedDateTime": "2023-07-15"
  }
},
{
  "resource": {
    "resourceType": "Observation",
    "id": "observation-1",
    "status": "final",
    "category": [
      {
        "coding": [
          {
            "system": "http://terminology.hl7.org/CodeSystem/observation-category",
            "code": "imaging"
          }
        ]
      }
    ],
    "code": {
      "coding": [
        {
          "system": "http://loinc.org",
          "code": "30746-1",
          "display": "CT of chest"
        }
      ]
    },
    "subject": {
      "reference": "Patient/patient-1"
    },
    "effectiveDateTime": "2023-07-10",
    "valueString": "Partial response to treatment"
  }
}
]
}
```

*Figure 4. Representational Example of FHIR-Compliant Profile Generated From Trained Large Language Model Using Free-Text Clinical Notes*

on large medical corpora capturing of the relevant knowledge, which helps in recognition and segmentation of specific terms with clinical notes. Syntactic and semantic parsing techniques also provide the ability to interpret relationships of entities with each other so that a complete data extraction can be achieved.

Once the structured data elements are extracted, they are meticulously mapped to corresponding FHIR resources. The patient resource captures demographic information, including identifiers, names, gender, birth date, and contact details, ensuring a complete patient profile. Clinical conditions and diagnoses are represented using the Condition resource, capturing essential details such as the onset date, verification status, and clinical status. The MedicationRequest resource is used to document prescribed medications, including the medication code, dosage, and administration instructions, while the Observation resource encodes clinical observations, lab results, and imaging findings. The Procedure resource documents clinical procedures and interventions, specifying the procedure type, date, and outcome, and the AllergyIntolerance resource records any known allergies and adverse reactions, including substance, reaction, and severity.

To ensure interoperability, the extracted data elements are mapped to standardized terminologies and coding systems. SNOMED CT is used for diagnoses and clinical findings, LOINC for lab results and observations, ICD-10 for disease classification, and RxNorm for medication identification. Integrating these standardized codes into the FHIR resources facilitates consistent interpretation and exchange of clinical information across different healthcare systems. This step involves cross-referencing the extracted terms with standardized vocabularies and ensuring accurate coding.

The individual FHIR resources are then combined into a FHIR Bundle, which encapsulates all related resources, presenting them as a cohesive patient profile. This "document" or "collection" bundle stores the patient's clinical information in a comprehensive and readily transferable way. The bundling process involves assembling the resources in a logical sequence, establishing relationships between them, and ensuring that all necessary references are correctly resolved.

### Restructuring FHIR Profiles for mCODE Compliant Data Structures

The primary step in effectively converting a FHIR profile into mCODE compliant data, and improving interoperability of oncology patients' data is to be extract data from the original FHIR profile. This information contains patient demographics, cancer diagnoses, treatments given (if any), observations made, and procedures performed. Some of these natural language processing (NLP) tools help in this extraction and to not miss any important clinical details. Then, the data is normalized from its original clinical coding systems and value sets to those required by mCODE (e.g., SNOMED CT for conditions, LOINC for laboratory tests, RxNorm for medications) [9].

After normalizing, data is assigned to the mCODE profiles. For example, patient demographic information will create the mCODE Patient profile but with a guarantee that all required identifiers and contact details are included in this structure. Cancer diagnoses are given as the mCODE CancerCondition profile, which contains clinical status and verification status along with diagnosis codes. Similarly, medications are mapped to the mCODE profile for medications, and observations or procedures will also map to their corresponding mCODE profiles. Following mapping, the mCODE profiles are rigorously vetted using tailored validation mechanisms to confirm implementation aligns with specifications laid out in mCODE. Integrity Test - This validates detailed integrity and accuracy of data to ensure that they are complete, correctly coded with the correct structure. The profiles are then reviewed by clinical and technical stakeholders for accuracy and relevance.

# Results and Validation of Large Language Model Outputs Towards

### FHIR and mCODE Profiles

This section presents the results of converting free-text clinical notes into structured FHIR profiles and transforming these profiles into mCODE-compliant formats. Using advanced natural language processing (NLP) techniques and large language models (LLMs), clinical information was accurately extracted from unstructured data. The subsequent transformation aligned these FHIR profiles with mCODE standards, adapting the data to fit within the oncology-specific framework.

This process involved rigorous validation to maintain data integrity and fidelity. Additionally we validate the accuracy of these results(FHIR and mCODE profiles) by placing them within open-source FHIR servers and tools like the FHIR R4 Validator for Conformance to determine how accurate our unique LLM-produced profiles are to those that are officially implemented within the industry [10]. The reasoning behind this method of validation is due to the intricacy of testing LLM accuracy without causing lapses in judgement for the LLM whilst creating profiles. Hence, exporting profiles to external serves and validators is the most effective way of measuring efficacy/accuracy from the model.

### Mapping and Processing Free-Text Data (e.g. Clinical Notes) to LLM's as FHIR Profiles

***Methodology of Pre-Processing Respective Free-Text Dataset:*** A multi-step, systematic approach using sophisticated natural language processing (NLP) and data structuring techniques was applied to enrich detailed FHIR profiles directly from the raw free-text clinical note for a leukemia patient. The clinical note (flow cytometry test details, specimen description and diagnostic interpretation) was first pre-processed meticulously. The purpose of this phase was to extract and classify the unstructured text, highlighting important medical concepts or entities. This included the extraction and categorization of terms such as "Mixed

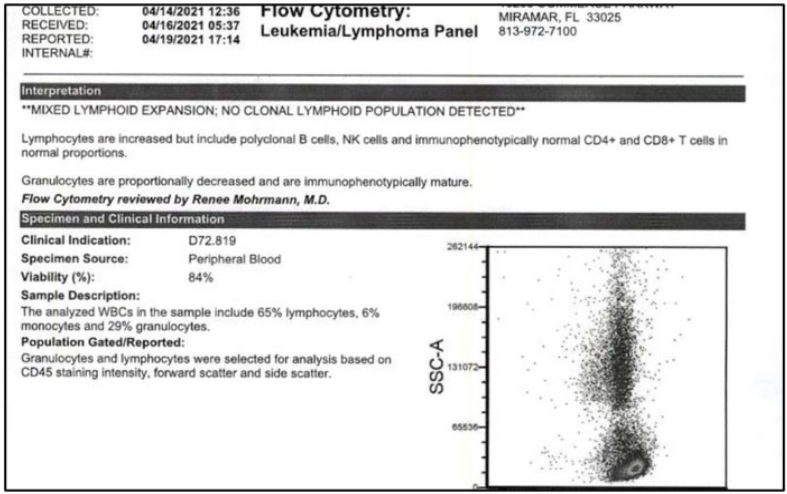

Figure 5. *Official Cancer Patient's Clinical Notes (Unstructured Data) Used to Breakdown Diagnosis Regarding Patient Status*

Lymphoid Expansion", "Peripheral Blood" and, "'84% viability'" by level of clinical relevance.

Following pre-processing, the next step involved mapping the extracted information to appropriate FHIR resources. The diagnostic report was translated into a `DiagnosticReport` resource, capturing the essential details such as the test type (Flow Cytometry), the clinical indication (D72.819), and the interpretation ("NO CLONAL LYMPHOID POPULATION DETECTED"). This resource was carefully structured to include relevant metadata, such as the collection and reporting dates, ensuring comprehensive documentation of the diagnostic process. The `Observation` resource was used to detail specific findings from the flow cytometry, such as the proportion of lymphocytes, granulocytes, and monocytes, ensuring each observation was accurately encoded with standard terminologies.

The `Specimen` resource was created to encapsulate the sample's details, including the specimen source (Peripheral Blood) and viability percentage (84%). This resource was enriched with precise information on the sample's composition, such as the proportions of lymphocytes, monocytes, and granulocytes. Each data point from the clinical note was meticulously mapped to corresponding fields within the `Specimen` resource, ensuring a high level of accuracy and completeness.

Furthermore, the process involved validating the structured FHIR profiles against HL7-FHIR standards to ensure they adhered to interoperability guidelines. This validation step was crucial in maintaining data integrity and ensuring the profiles could be seamlessly integrated into broader healthcare systems.

## Mapping/Processing Generated FHIR Profiles to LLMs for mCODE Conversion

We initiated our process with a comprehensive mapping between FHIR resources and mCODE elements, guided by the mCODE Implementation Guide. This mapping identified direct correspondences and areas requiring extension. For instance, in translating a DiagnosticReport resource to the mCODE cancer-diagnostic-report profile, we meticulously aligned elements such as status, category, and code with mCODE specifications. The category coding for Hematology" (HM) and the LOINC code "55233-1" for the Leukemia/Lymphoma Panel were precisely mapped to ensure semantic consistency.

For oncology-specific concepts not natively represented in FHIR, we developed custom extensions. These extensions were designed to capture the granularity required by mCODE while adhering to FHIR's extensibility principles. In the case of genomic data elements, crucial for comprehensive oncological profiling, we created extensions that could encapsulate

```
{
  "resourceType": "DiagnosticReport",
  "id": "leukemia-lymphoma-panel-20210419",
  "status": "final",
  "category": {
    "coding": [
      {
        "system": "http://terminology.hl7.org/CodeSystem/v2-0074",
        "code": "HM",
        "display": "Hematology"
      }
    ]
  },
  "code": {
    "coding": [
      {
        "system": "http://loinc.org",
        "code": "55233-1",
        "display": "Leukemia/Lymphoma Panel"
      }
    ],
    "text": "Leukemia/Lymphoma Panel"
  },
  "subject": {
    "reference": "Patient/12345"
  },
  "effectiveDateTime": "2021-04-14T12:36:00Z",
  "issued": "2021-04-19T17:14:00Z",
  "performer": [
    {
      "actor": {
        "display": "Renee Mohrmann, M.D."
      }
    }
  ],
  "specimen": [
    {
      "reference": "Specimen/1"
    }
  ],
  "result": [
    {
      "reference": "Observation/lymphoid-expansion"
    },
    {
      "reference": "Observation/sample-description"
    }
  ]
}
```

```
{
  "resourceType": "Observation",
  "id": "lymphoid-expansion",
  "status": "final",
  "category": [
    {
      "coding": [
        {
          "system": "http://terminology.hl7.org/CodeSystem/observation-category",
          "code": "laboratory",
          "display": "Laboratory"
        }
      ]
    }
  ],
  "code": {
    "coding": [
      {
        "system": "http://loinc.org",
        "code": "55233-1",
        "display": "Leukemia/Lymphoma Panel"
      }
    ],
    "text": "Mixed Lymphoid Expansion"
  },
  "subject": {
    "reference": "Patient/12345"
  },
  "effectiveDateTime": "2021-04-14T12:36:00Z",
  "valueString": "NO CLONAL LYMPHOID POPULATION DETECTED. Lymphocytes are increased but include polyclonal B cells, NK cells, and immunophenotypically normal CD4+ and CD8+ T cells in normal proportions. Granulocytes are proportionally decreased and are immunophenotypically mature."
}
```

**Figure 6a & 6b.**
*Official Large Language Model Generated Diagnostic Report and Observation Profile Respective to Cancer Patient's Clinical Notes/Free-Text Data in Fig. 5*

complex information such as biomarker status and genomic variants. This approach allowed us to represent vital data that might be derived from the leukemia/lymphoma panel results, ensuring it was accurately captured within the mCODE framework [11].

Our process also accounted for the temporal and provenance aspects crucial in oncology care. The effectiveDateTime and issued fields in the DiagnosticReport were carefully preserved in the mCODE profile, maintaining the critical timeline of diagnostic events. Furthermore, we ensured that references to related resources, such as the performing clinician, the patient, and associated specimens and observations, were accurately translated to maintain the integrity of the clinical narrative within the mCODE structure.

**Training and Evaluation of Accuracy Rates via Large Language Models**

The provided dataset is instrumental in training a language model (LLM) to accurately generate FHIR and mCODE profiles. When analyz-ing and utilizing results provided form an LLM, it's imperative to determine the results abide to highly-measured accuracy rates. This starts, by providing our LLM with over 70 patient profiles and converting each one to both FHIR and mCODE compliant formats. Whilst doing this, our LLM monitored and stored patterns to continue to look for when both analyzing free-text data such as clinical notes, but also in terms of producing patient profiles.

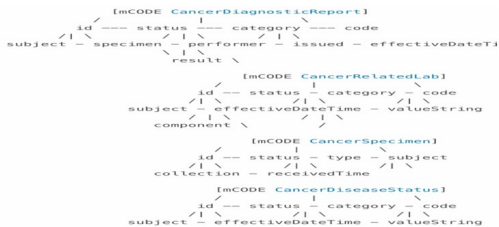

*Figure 7.*
mCODE Spider Diagram to Demonstrate Framework/Breakdown for Data Elements Derived from FHIR Profiles and Variables

| Variable | Source | Values | Source of Values |
|---|---|---|---|
| cancer diagnosis | PDF progress notes | Anal Cancer, Bone/Soft Tissue Cancer, Brain/CNS Cancer, Breast Cancer, Cervical Cancer, Colon/Colorectal Cancer, Endocrine Cancer - pituitary, thyroid, parathyroid, Endometrial Cancer, Esophageal/Esophagogastric Junction Cancer, Gastric Cancer - Stomach, Gynecological Other or Vaginal Cancer, Head/Neck Cancer, Hematologic | Master CRF - Additional New Primary |
| cancer diagnosis date | PDF progress notes | mm/dd/yyyy | N/A |
| metastasis indication | PDF progress notes | Yes, No | N/A |
| metastasis site | PDF progress notes | Brain/Central Nervous System (CNS), Head/Neck, Breast, Lung, Chest/Thorax - Other, Liver, Abdomen - Other, Pelvis, Bone(s), Lymph Node(s) - Distant, Lymph Node(s) - Local, Lymph Node(s) - Unspecified, Fluid - Pleural/Abdominal, Muscle(s), Skin, Soft Tissue, Other, Not Documented | Master CRF - Staging |
| stage - TNM T | PDF | Tx, T0, Tis, T1, T1mi, T1a, T1a1, T1a2, T1b, T1b1, T1b2, T1b3, T1c, T1c1, T1c2, T1c3, T2, T2a, T2a1, T2a2, T2b, T2c, T3, T3a, T3b, T3c, T4, T4a, T4b, T4c, T4d, Other, Not Documented | Master CRF - Staging |
| | progress notes | N/A (NER model, needs to be standardized to PDF values listed above) | |
| stage - TNM N | PDF | NX, N0, N0(i+), N0(mol+), N0a, N0b, N1, N1mi, N1a, N1b, N1c, N2, N2mi, N2a, N2b, N2c, N3, N3a, N3b, N3c, Other, Not Documented | Master CRF - Staging |
| | progress notes | N/A (NER model, needs to be standardized to PDF values listed above) | |
| stage - TNM M | PDF | MX, M0, M0(i+), M1, M1a, M1a(0), M1a(1), M1b, M1b(0), M1b(1), M1c, M1c(0), M1c(1), M1d, M1d(0), M1d(1), Other, Not Documented | Master CRF - Staging |
| | progress notes | N/A (NER model, needs to be standardized to PDF values listed above) | |
| stage - numerical | PDF | I, IA, IA1, IA2, IA3, IB, IB1, IB2, IB3, IC, IC1, IC2, IC3, II, IIA, IIA1, IIA2, IIB, IIC, III, IIIA, IIIA1, IIIA1i, IIIA1ii, IIIA2, IIIB, IIIC, IIIC1, IIIC2, IIID, IV, IVA, IVB, IVC, Other, Not Documented | Master CRF - Staging |
| | progress notes | N/A (NER model, needs to be standardized to PDF values listed above) | |
| Histology | PDF progress notes | Acinar cell carcinoma, Acral/Acral lentigous, Adenocarcinoma, Adenocarcinoma - Diffuse type, Adenocarcinoma - Intestinal type, Adenocarcinoma – Mucinous, Adenocarcinoma – Papillary, Adenocarcinoma – Tubular, Adenoid cystic carcinoma, Adenosquamous carcinoma, Astrocytoma, Basal cell carcinoma, Brenner tumor, | Master CRF - Initial Diagnosis |
| Histology Grade | PDF progress notes | G1 (low grade; well differentiated), G2 (intermediate grade; moderately differentiated), G3 (high grade; poorly differentiated), G4 (high grade; undifferentiated), GX-cannot be assessed (Undetermined), Not Documented | Master CRF - Initial Diagnosis |
| Cancer Laterality | PDF progress notes | Unilateral - Left, Unilateral - Right, Bilateral, Not Documented | Master CRF - Initial Diagnosis |

*Figure 8. Table of Monitored and Charted Patterns Captured from Large Language Model Derived from Thousands of Unstructured Raw Datasets*

By incorporating a variety of clinical variables such as cancer types, diagnosis dates, metastasis indications, staging information, histology grades, and laterality, the dataset allows the LLM to understand and classify these variables with precision. This classification process is crucial for ensuring that the generated profiles accurately reflect a patient's clinical data. For instance, the LLM learns that "Bone/Soft Tissue Cancer" corresponds to a specific cancer diagnosis and that "Head/Neck" represents a metastasis site, enabling it to populate the correct FHIR elements such as Condition. code and Observation.value Codeable Concept.

Mapping these clinical variables to FHIR elements is a significant step facilitated by the dataset. Each entry guides the LLM on how to structure data within FHIR profiles. For example, staging information like "TX, T0, Tis, T1" is mapped to FHIR Observation. code and Observation. value Codeable Concept, ensuring consistency and accuracy. The dataset's standardized terminologies and values, sourced from master CRFs, help the LLM maintain compatibility with global healthcare standards. This standardization is vital for generating both FHIR and mCODE profiles, as it ensures that elements such as cancer diagnosis dates (Observation.effectiveDateTime) and metastasis indications (Condition.extension) are accurately encoded.

The structured nature of the dataset also plays a critical role in mitigating errors. By providing clear delineations of variables and their values, the dataset

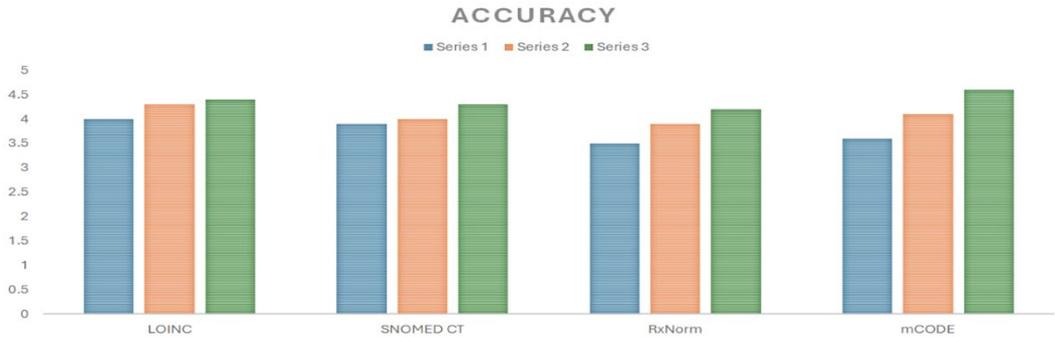

*Figure 9. Charted Data to Visualize Large Language Model Accuracy Rates in Determining Respective Digital Resources per Dataset of Thousands of Profiles*

reduces ambiguity and enhances the LLM's accuracy in generating profiles. This structured approach is particularly beneficial when dealing with data from multiple sources, such as PDFs and progress notes, as it allows the LLM to extract and standardize relevant information effectively.

**Accuracy Rates of LLM-Generated mCODE Resources within FHIR Profiles – A Trial versus Trial Comparison**

The model predicted the correct SNOMED-CT, LOINC code in 87% of the raw data and 90% of the standard data by test frequency and around 84% for RxNorm. In the subset of raw data where the original and model-predicted LOINC codes disagreed, the model-predicted LOINC code was correct in 85% of the data by test frequency. There is an opportunity to improve around generic medication and raw results and clinical data by training the model with smaller data set. mCODE conformance improved in Series 3 with profile interoperability compliance coming around 92%.

Additionally, it is important to draw upon the observation that throughout the 3 separate trials(series) performed via our Large Language Model, accuracy rates continued to see a constant increase. However, regardless of the digital recourse being measured, accuracy rates saw an increase between trials, hence demonstrating our model learning from itself. However, it is also important to look at the larger impact.

The impact of these accuracy rates on clinical practice and research is significant. High accuracy in coding ensures that healthcare providers have access to precise and standardized information, which is vital for patient safety, accurate diagnosis, and effective treatment planning. Improved mCODE conformance and profile interoperability compliance at 92% indicate that our model supports oncology data standards effectively, facilitating better data sharing and integration in cancer care. This revolutionary accuracy, completely redefines the status quo with other LLM's such as GPT-4 and Claude averaging between 77-80%. This high level of conformance is particularly important for advancing precision medicine and enabling researchers to draw more accurate insights from aggregated health data [12].

# Development of an Integrated Clinical Trial Matching Engine for Oncology Patients Using LLM Produced FHIR Resources and mCODE Profiles

The integration of clinical trial matching services with electronic health records (EHRs) and patient data repositories leverages advanced data standards like FHIR and specialized oncology profiles such as mCODE. This integration holds the potential to significantly enhance the precision and efficiency of matching patients to appropriate clinical trials, thereby optimizing therapeutic outcomes and advancing personalized medicine.

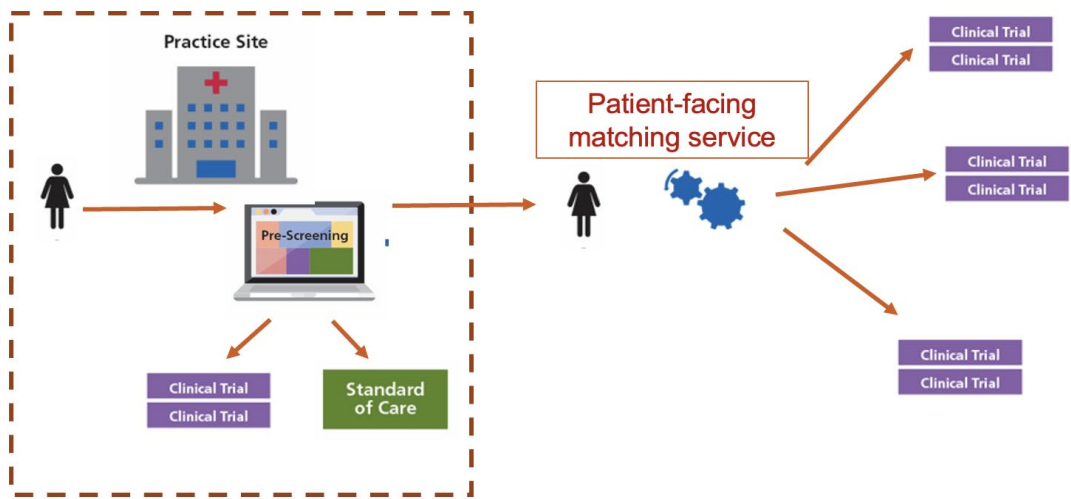

*Figure 10. Visualized Framework for Integrated Trial Matching Service/Engine for All Hospitalized Cancer Patients with mCODE Profiles Produced via LLM Model*

In the context of oncology, clinical trial matching is particularly crucial. Cancer treatment often requires highly personalized approaches due to the variability in tumor biology and patient response to therapies. Traditional methods of clinical trial enrollment are frequently manual, time-consuming, and reliant on limited patient data, which can result in missed opportunities for patients to participate in trials that might offer them the best possible treatment. By utilizing mCODE, a standardized data model that captures essential oncology data elements, alongside FHIR, which facilitates the seamless exchange of healthcare information, we can automate and streamline this matching process. This ensures that comprehensive and precise patient data is readily available for matching algorithms, leading to more accurate identification of suitable clinical trials for patients.

Hence, the use of mCODE profiles as mentioned throughout the earlier portions of this paper, ensures that critical cancer-specific data elements such as tumor characteristics, treatment history, genomic information, and patient outcomes are consistently captured and structured. When this standardized data is made interoperable through FHIR, it can be easily shared across various platforms and systems, including clinical trial databases. This interoperability not only accelerates the matching process but also improves the accuracy and relevance of matches. Patients can be matched to trials that are tailored to their unique clinical and genetic profiles, thus increasing the likelihood of trial success and contributing to the broader goals of precision oncology.

## Development of a Framework for Proactive and Efficient Integrated Clinical Trial Matching for Cancer Patients

The provided framework represents a comprehensive and systematic approach to leveraging mCODE FHIR profiles for clinical trial matching, underscoring its feasibility and integration with existing healthcare systems. This framework starts with the extraction of detailed patient treatment data, encapsulated within mCODE profiles, which include expanded data elements pertinent to oncology. The mCODE profiles are derived from standard mCODE profiles, enriched as needed to ensure all relevant patient information is captured accurately – this process was demonstrated above in the methodology of translating profiles from free-text data into FHIR mCODE profiles. This enriched data serves as a robust foundation for identifying clinical trial matches, ensuring that the patient's unique clinical characteristics and treatment history are fully considered.

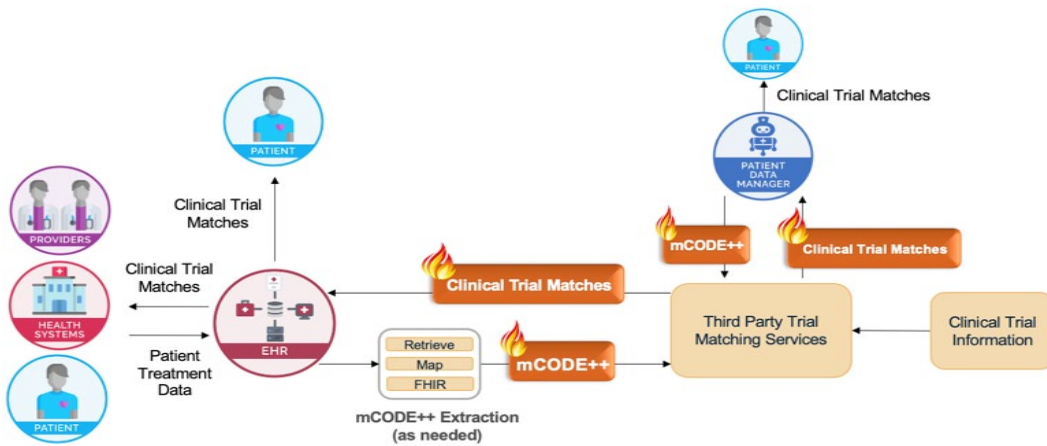

*Figure 11. Detailed Framework for Breakdown of Integrated Clinical Trial Matching Service/Engine and Interoperability Across Platforms/Providers*

In the framework, patient treatment data is first retrieved and transformed into mCODE profiles. These profiles, adhering to FHIR standards, facilitate seamless data exchange and interoperability, a critical aspect for integrating with third-party trial matching services. The use of FHIR ensures that the data is in a standardized format, making it compatible with various health information systems and databases. Once the mCODE++ profiles are prepared, they are transmitted to clinical trial matching algorithms, which analyze the profiles against a database of available clinical trials. This step is crucial as it leverages advanced computational methods to identify the most suitable trials for each patient based on their specific clinical attributes.

The framework's integration with third-party trial matching services further enhances its utility. These services utilize the mCODE profiles to generate linked trial matches, providing detailed information on relevant clinical trials that align with the patient's profile. Our third-party clinical trial matching engine by using a SMART on FHIR application (FHIR guidelines for apps) that sends relevant patient data derived from mCODE and FHIR profiles and returns uniform results [13]. Our specific clinical trial matching engine utilized a backend library that was constructed to provide a generic shell for services to connect to a behind-the-scenes trial matching service and engine. To accomplish such, our matching engine receives a FHIR bundle of resources as provided from the FHIR and mCODE profiles generated above in sections 1-3. Our engine then consequently uses those resources derived from the profiles to generate a FHIR search result that contains FHIR ResearchStudy(similar to any other variable/object that comprises a FHIR and mCODE profile) objects that describe matching clinical trials. However, on the other hand, now that we have discussed how the back-end works, let's discuss the imperative front-end services.

In regard to the front-end services of our clinical trial matching engine, we use a TrialScope Wrapper that calls TrialScope's API for the front end of clinical matching service, to help patients, hospitals, and doctors interact with the service by inputting their own profiles, customizing and filtering services, etc. [14]. As previously mentioned, our backend services break down the FHIR profile, similar to a reverse methodology of our free-text data to mCODE conversion.

*Figure 12. Screenshot of Official Interface of Integrated Clinical Trial Matching Service/Engine to Demonstrate All Possible Trials Based on Patient Details*

Because of this, our service breaks down the FHIR profile back into free-text data and variables, however, to implement a mechanism to find associated clinical trials with these profiles we rely on a third-party server, namely – BreastCancerTrials. This third-party tool is integrated into our system to be able to send queries from our service to the breastcancertrials.org clinical trial search service [15].

The framework supports the continuous flow of information, where clinical trial matches are fed back into the system, allowing healthcare providers to present patients with up-to-date and precise trial options. This iterative process not only enhances the accuracy of trial matching but also ensures that patients are consistently presented with the best possible treatment opportunities. The feasibility and effectiveness of this framework lie in its adherence to FHIR standards and the comprehensive nature of mCODE profiles.

### Demonstration of Integrated Clinical Trial Matching Service and Respective Breakdown of Engine Results/Trial Outputs

The integrated clinical trial matching service leverages mCODE FHIR profiles to deliver highly personalized and relevant clinical trial matches for patients, particularly those battling cancer. This is exemplified in the provided screenshot, which shows a breast cancer patient matched with a clinical trial studying the molecular mechanisms of clinical resistance to targeted therapy. This matching process is immensely beneficial for patients, as it provides them with opportunities to participate in cutting-edge research that might offer new treatment options when and where standard therapies have failed. The likelihood of a match is determined by analyzing detailed patient data encapsulated in mCODE profiles, ensuring that the trials they are matched with are highly pertinent to their specific clinical characteristics and needs.

For cancer patients, this system can be a lifeline. The matching service not only identifies suitable trials but also streamlines the process of enrollment, allowing patients to access potentially life-saving treatments more quickly. The trial highlighted in the screenshot below, conducted by the Memorial Sloan Kettering Cancer Center, aims to understand why tumors become resistant to therapy. Such research is crucial for developing new strategies to overcome resistance and improve patient outcomes. By participating in their own respective studies and clinical trials patients contribute to valuable scientific knowledge while also receiving cutting-edge care that may offer

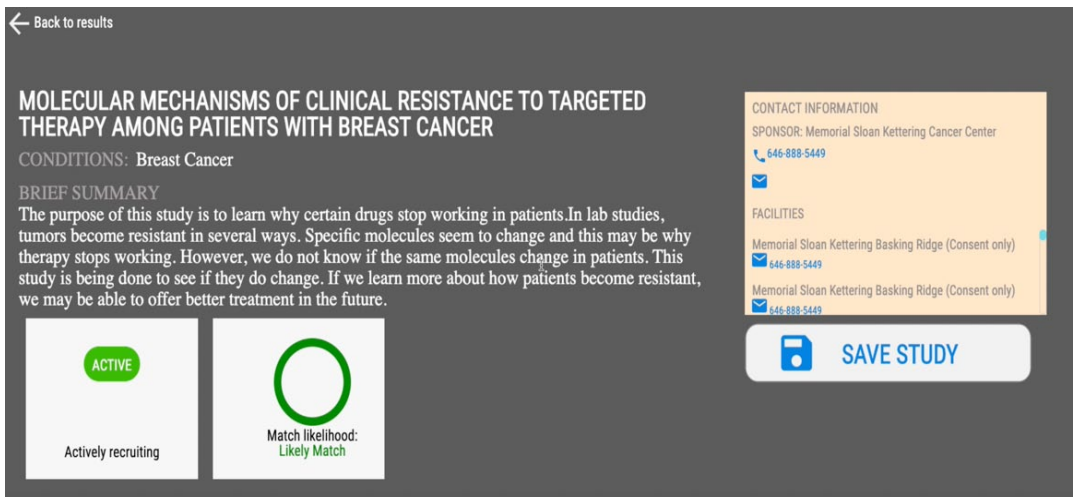

*Figure 13. Screenshot of Official Interface of Integrated Clinical Trial Matching Service/Engine to Demonstrate Details of Selected/Specific Trial for Patient*

better therapeutic results compared to existing treatment options.

Healthcare providers and researchers benefit significantly from this integrated trial matching service. It enhances the efficiency of clinical trial recruitment, ensuring that eligible patients are quickly and accurately identified. This reduces the time, and resources spent on finding suitable participants, allowing researchers to focus more on conducting the trial and less on recruitment logistics. Additionally, the standardized nature of mCODE FHIR profiles ensures interoperability across various health information systems, facilitating seamless data exchange and collaboration. This integrated approach not only accelerates the pace of clinical research but also improves the quality and relevance of trial outcomes, ultimately leading to advancements in cancer treatment and patient care.

*Figure 14. Screenshot of Official Interface of Integrated Clinical Trial Matching Service/Engine to Demonstrate Eligibility of Trial Cohorts for Patient Info*

# Discussion and Final Conclusions

Our research leverages HL7-FHIR (Fast Healthcare Interoperability Resources) to address significant challenges in healthcare data interoperability. In this study, we implemented FHIR to streamline data exchange processes and enhance the integration of electronic health records (EHRs) across diverse healthcare systems. This implementation facilitated seamless communication between disparate systems, enabling accurate sharing of patient information [16].

A key aspect of our work involved the use of FHIR's resource-based approach to manage patient data. Each piece of information, such as allergy lists and medication histories, was treated as a discrete resource. This modular approach enabled the software to provide a very structured means of sharing data and prevented data mismatch or errors. In the case of the present research, this capability was essential in designing a system which would allow for changes in clinical practice while not requiring extensive modifications. Moreover, they implemented FHIR using the modern web technologies such as RESTful API, JSON and XML.

Concerning patient centricity, the current study also sought to establish what FHIR meant for patient orientation. We created patient's portals and applications that applied FHIR to enhance the patients' understanding of their data. This empowered patients to transfer their information to various care givers hence integration of care. For instance, patient migration meant that records could easily be transferred to new caregivers thus improving theMedical history of a patient. By developing this portal, this paper revolutionized access to custom care for cancer patients around the world. Achieving a 92% accraucy rate in profile accuracy, this paper trumps current solutions in the status quo and paves a new path for cancer care around the world.

Further, this explored the application of FHIR in the enhancement of other progressive forms of healthcare like clinical decision support systems (CDSS) and precision medications. Based on precision medicine, we explained how FHIR could help reconcile genomic information with conventional clinical data for tailored treatment plans based on a patient's genetic makeup.

Future research opportunities for follow-up research based on this work are the development of of artificial intelligence (AI) and machine learning (ML) based application on FHIR. The normalization of data for feeding to AI models can result into delivering of analytical tools for early warning of likely patient complications and direction of resources towards more likely complications among the patient population. Also, understanding FHIR capabilities regarding enhancing international interoperability of the healthcare system is highly important regarding the contemporary threats that affect global health such as pandemics.

In conclusion, by utilizing FHIR's standardized protocols and modular architecture, we achieved significant improvements in data exchange, patient-centric care, and advanced healthcare applications. Our findings underscore the potential of FHIR to transform healthcare systems, and we encourage further research to explore and expand its capabilities, fostering a more connected, efficient, and innovative healthcare ecosystem [17].

# References/Citations

Liu, C., Liu, X., Wu, F., Xie, M., Feng, Y., & Hu, C. (2018). Using Artificial Intelligence (Watson for Oncology) for Treatment Recommendations Amongst Chinese Patients with Lung Cancer: Feasibility Study. *JMIR. Journal of Medical Internet Research/Journal of Medical Internet Research*, *20*(9), e11087. https://doi.org/10.2196/11087

*How Apple's Health Records is Reshaping Patient Engagement at Johns Hopkins Medicine*. (2020, April 30). NGPX 2024. https://patientexperience.wbresearch.com/blog/apple-health-record-strategy-reshaping-patient-engagement-at-johns-hopkins-medicine-clinic

Cascella, M., Semeraro, F., Montomoli, J., Bellini, V., Piazza, O., & Bignami, E. (2024). The Breakthrough of Large Language Models Release for Medical Applications: 1-Year Timeline and Perspectives. *Journal of Medical Systems*, *48*(1). https://doi.org/10.1007/s10916-024-02045-3

Mandel, J. C., Kreda, D. A., Mandl, K. D., Kohane, I. S., & Ramoni, R. B. (2016). SMART on FHIR: a standards-based, interoperable apps platform for electronic health records. *Journal of the American Medical Informatics Association*, *23*(5), 899–908. https://doi.org/10.1093/jamia/ocv189

Osterman, T. J., Terry, M., & Miller, R. S. (2020). Improving cancer data interoperability: The promise of the Minimal Common Oncology Data Elements (MCODE) initiative. *JCO Clinical Cancer Informatics*, *4*, 993–1001. https://doi.org/10.1200/cci.20.00059

Post, A. R., Burningham, Z., & Halwani, A. S. (2022). Electronic Health Record Data in Cancer Learning Health Systems: Challenges and opportunities. *JCO Clinical Cancer Informatics*, *6*. https://doi.org/10.1200/cci.21.00158

Johnson, S. B., Bakken, S., Dine, D., Hyun, S., Mendonça, E., Morrison, F., Bright, T., Van Vleck, T., Wrenn, J., & Stetson, P. (2008). An electronic health record based on structured narrative. *Journal of the American Medical Informatics Association*, *15*(1), 54–64. https://doi.org/10.1197/jamia.m2131

Wen, A., Rasmussen, L. V., Stone, D., Liu, S., Kiefer, R., Adekkanattu, P., Brandt, P. S., Pacheco, J. A., Luo, Y., Wang, F., Pathak, J., Liu, H., & Jiang, G. (2021). *CQL4NLP: Development and integration of FHIR NLP extensions in Clinical Quality language for EHR-driven phenotyping*. PubMed Central (PMC). https://www.ncbi.nlm.nih.gov/pmc/articles/PMC8378647/

*SNOMED CT, LOINC, and ICD-10 – the foundations of semantic interoperability*. (n.d.). Wemedoo. https://wemedoo.com/snomed-ct-loinc-and-icd-10/

*How to validate Fhir R4 Resource based on profiles and custom structure definition?* (n.d.). Stack Overflow. https://stackoverflow.com/questions/58056197/ho-to-validate-fhir-r4-resource-based-on-profiles-and-custom-structure-definition

Svoboda, M., Behulova, R. L., Slamka, T., Sebest, L., & Repiska, V. (2023). Comprehensive genomic profiling in predictive testing of cancer. *Physiological Research*, *72*(S3), S267–S275. https://doi.org/10.33549/physiolres.935154

*Introducing MCODEGPT: A Revolutionary step in cancer data standardization and Sharing*. (2023, September 14). OHDSI Forums. https://forums.ohdsi.org/t/introducing-mcodegpt-a-revolutionary-step-in-cancer-data-standardization-and-sharing/19828

Jin, Q., Wang, Z., Floudas, C. S., Chen, F., Gong, C., Bracken-Clarke, D., Xue, E., Yang, Y., Sun, J., & Lu, Z. (n.d.). *Matching Patients to Clinical Trials with Large Language Models*. PubMed Central (PMC). https://www.ncbi.nlm.nih.gov/pmc/articles/PMC10418514/

*TrialScope Engage™ – TrialScope*. (n.d.). https://www.trialscope.com/trialscope-engage/

*Find Breast Cancer Trials w/ BCT's Matching Tool*. (n.d.). https://www.breastcancertrials.org/BCTIncludes/FindATrial/GetStarted.html

Gazzarata, R., Almeida, J., Lindsköld, L., Cangioli, G., Gaeta, E., Fico, G., & Chronaki, C. E. (2024). HL7 Fast healthcare interoperability resources (HL7 FHIR) in digital healthcare ecosystems for chronic disease Management: Scoping review. *International Journal of Medical Informatics*, *189*, 105507. https://doi.org/10.1016/j.ijmedinf.2024.105507

McLelland, R. (2024, July 6). *Unlocking the Potential of FHIR: An Overview of its Impact and Future*. UHIN. https://uhin.org/blog/unlocking-potential-of-fhir/